\pdfoutput=1

\documentclass[11pt]{article}

\usepackage[preprint]{acl}

\usepackage{times}
\usepackage{latexsym}
\usepackage{amssymb}
\usepackage[T1]{fontenc}

\usepackage[utf8]{inputenc}

\usepackage{microtype}

\usepackage{inconsolata}

\usepackage{graphicx}
\usepackage{color, colortbl}
\definecolor{Gray}{gray}{0.95}
\usepackage{multirow}
\usepackage{tablefootnote}
\title{CDChat: A Large Multimodal Model for Remote Sensing Change Description}

\author{
  \textbf{Mubashir Noman\textsuperscript{1}},
  \textbf{Noor Ahsan\textsuperscript{1}},
  \textbf{Muzammal Naseer\textsuperscript{2}},
  \textbf{Hisham Cholakkal\textsuperscript{1}},
\\
  \textbf{Rao Muhammad Anwer\textsuperscript{1}},
  \textbf{Salman Khan\textsuperscript{1,3}},
  \textbf{Fahad Shahbaz Khan\textsuperscript{1,4}}
\\
  \textsuperscript{1}MBZUAI U.A.E,
  \textsuperscript{2}Khalifa University U.A.E,
  \textsuperscript{3}Australian National University, Australia,
\\
  \textsuperscript{4}Linköping University, Sweden
}

\begin{document}
\maketitle
\begin{abstract}
Large multimodal models (LMMs) have shown encouraging performance in the natural image domain using visual instruction tuning. However, these LMMs struggle to describe the content of remote sensing (RS) images for tasks such as image or region grounding, classification, etc. Recently, GeoChat make an effort to describe the contents of the RS images. Although, GeoChat achieves promising performance for various RS tasks, it struggles to describe the changes between bi-temporal RS images which is a key RS task. This necessitates the development of an LMM that can describe the changes between the bi-temporal RS images. However, there is insufficiency of datasets that can be utilized to tune LMMs. In order to achieve this, we introduce a change description instruction dataset that can be utilized to finetune an LMM and provide better change descriptions for RS images. Furthermore, we show that the LLaVA-1.5 model, with slight modifications, can be finetuned on the change description instruction dataset and achieve favorably better performance. Code and models are available at \url{https://github.com/techmn/cdchat}.
\end{abstract}

\section{Introduction}
\label{sec:intro}

Recent progress in the large multimodal models (LMMs) \citep{liu2023improvedllava, openai2024gpt4} has urged the researchers to utilize it for various vision application domains such as remote sensing \citep{kuckreja2023geochat}, medical imaging \citep{Omkar2023XrayGPT}, etc.
These LMMs serve as general purpose assistants and demonstrate impressive performance on various tasks like image grounding, scene classification, visual question answering (VQA), etc.
Subsequently, \citet{kuckreja2023geochat} demonstrated the ability of LMMs in remote sensing (RS) field and introduced the GeoChat model that can perform various conversational tasks. 
However, GeoChat strives in describing the semantic changes between the co-registered satellite image pair. 
As RS domain lacks the multi-modal conversational data for instruction-tuning, therefore, \citet{kuckreja2023geochat} prepared the conversational dataset by utilizing the existing RS datasets of scene classification and object detection. This aided the GeoChat model to improve its performance for RS imagery. The GeoChat performs the visual instruction-tuning of LMM on RS data comprising of single image and text pairs. 
However, RS change description task requires co-registered image pairs along with the text descriptions that narrates the changes between them. In RS domain, change detection (CD) refers to identifying the semantic changes between the co-registered bi-temporal RS images. Similar to the other RS datasets, RS domain also lacks the conversational datasets for change detection task and it requires strenuous effort to manually annotate the RS image pairs and get corresponding image-text pairs. 
To this end, we attempt to create a conversational change description dataset that can be utilized for instruction-tuning of LMM and improves performance of the LMM for RS change description task.

\begin{figure}[t]
  \includegraphics[width=\columnwidth]{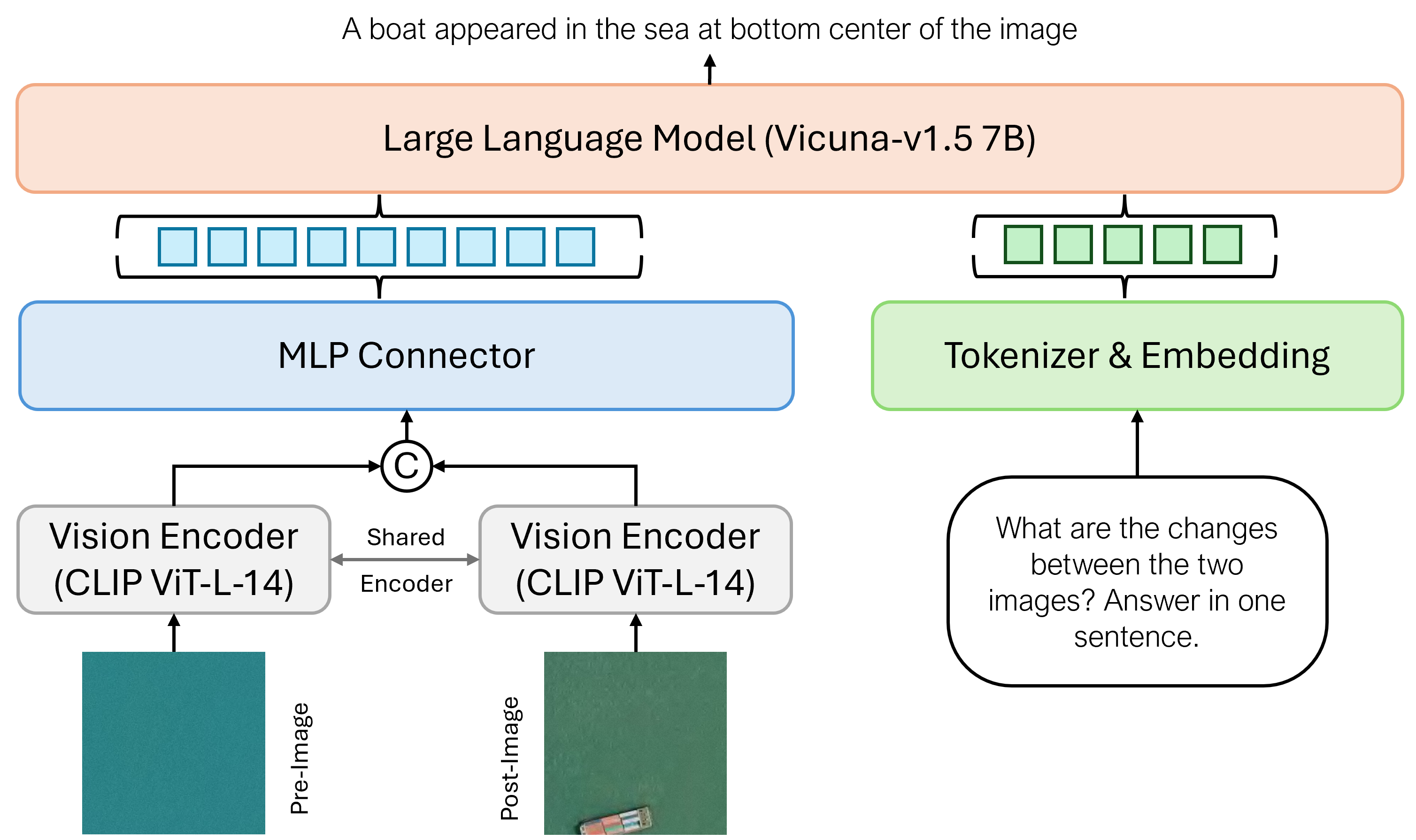}
  \caption{An overview of the CDChat. It comprises of shared vision encoder (ViT-L-14) to extract bi-temporal image features, MLP connector to project the image features to language space, and an LLM to generate the query response.}
  \label{fig:cdchat_architecture}
\end{figure}

In this paper, we propose CDChat that is a conversational assistant for RS change description task. We manually annotate the SYSU-CD \citep{shi21sysucd} dataset to obtain the change text and image pairs. Similar to other works, we utilize Vicuna-v1.5 \citep{vicuna2023} to generate the instruction data comprising 19k conversations. We create change text and image pairs from the two large scale change detection datasets including SYSU-CD \citep{shi21sysucd} and LEVIR-CD \cite{chen2020_levircd}. Specifically, we generate the multi-round VQA pairs that are related to describing the change regions in the image as well as counting the number of change regions. To summarize, our contributions are under:
\begin{itemize}
    \item We manually annotate the SYSU-CD \citep{shi21sysucd} dataset to obtain the text descriptions of the changes between the bi-temporal RS images. Utilizing the segmentation masks, we calculate the number of change regions present within the bi-temporal image pairs.
    \item We generate the instructional dataset for VQA change detection task by utilizing Vicuna-v1.5 \citep{vicuna2023} within an automated pipeline.
    \item We perform the low rank adaptation (LORA) \citep{hu2022lora} finetuning of LLaVA-1.5 \citep{liu2023improvedllava} model by employing our instructional change description dataset for RS change description task referred as CDChat. We demonstrate that the CDChat performs better compared to the existing LMMs.
\end{itemize}

\section{Annotation of CD Datasets}
\begin{figure}[t]
  \includegraphics[width=\columnwidth]{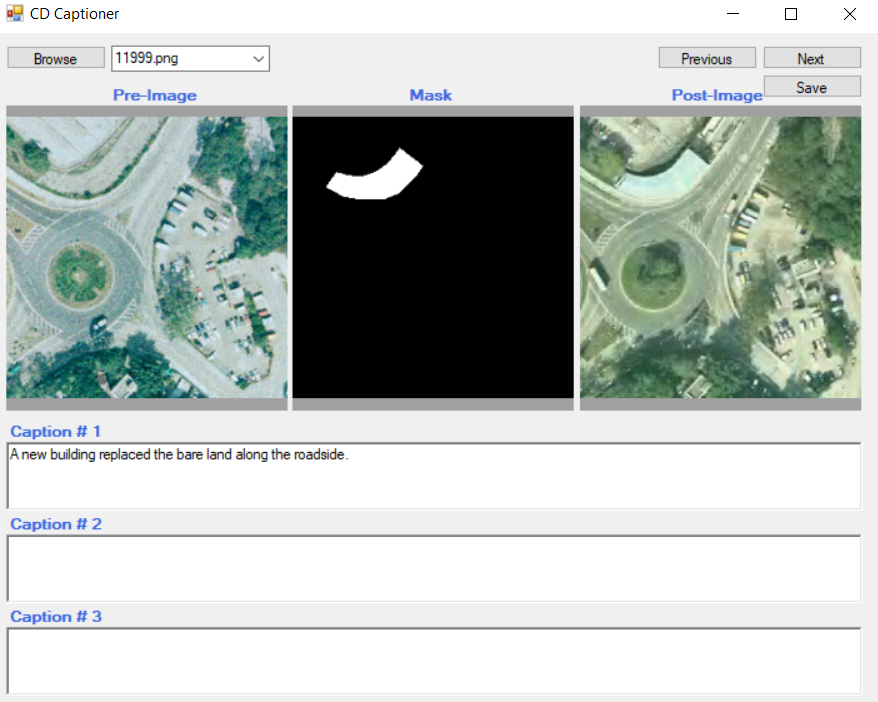}
  \caption{A custom graphical user interface developed for annotation of SYSU-CD \citep{shi21sysucd} dataset. The tool allows the annotator to write the change captions for the image pair by looking into the pre and post-change images along with the corresponding segmentation mask.}
  \label{fig:cdchat_annotation_tool}
\end{figure}

Existing RS change detection (CD) datasets mainly focus on the changes related to building construction and demolition. However, SYSU-CD is a large scale public CD dataset that provides the segmentation masks for changes related to building construction, ground work before construction, sea construction, road expansion, and vegetation changes.
We therefore selected the SYSU-CD for annotation purpose. We created a custom graphical user interface (GUI) tool to generate the text descriptions from the bi-temporal images and segmentation masks. Figure.~\ref{fig:cdchat_annotation_tool} shows the screenshot of the GUI tool used for annotation purpose. 
The tool allowed the annotators to look at the change masks and write multiple descriptions about the change regions. The GUI tool was enabled to set back and forth between the image pairs by using keyboard for easy access and fast annotation process.
A team of graduate students is composed to produce the change text descriptions. The annotated change descriptions are verified by verification team before utilizing it for instruction dataset generation.
After generating the text descriptions, we utilize the OpenCV \citep{opencv_library} library to find the number of change regions within a segmentation mask. This information of change region count is combined with the annotated change descriptions to obtain the final text descriptions for each bi-temporal image pair.

Besides, LEVIR-CC \citep{liu2022_levircc} dataset provide the change captions for the LEVIR-CD \citep{chen2020_levircd} dataset. However, it omits the segmentation masks of the image pairs. We match the change captions of the LEVIR-CC with the ground truth masks of the LEVIR-CD and combine it with the annotated dataset to increase the dataset size. 

\subsection{CD Instruction Dataset}
\label{sec:instruction_dataset}
To generate the multi-round conversation dataset, we utilize the Vicuna-v1.5(7B) \citep{vicuna2023} model. We provide system instructions and change descriptions to Vicuna and ask to generate a conversation in a manner like it is visualizing the bi-temporal images. To generate high quality question-answer pairs from the change descriptions, we provide few-shot examples to the Vicuna model as further instructions. In particular, we are able to generate approximately 19k multi-round conversations from the two large scale public CD datasets.

\section{Approach}
\label{sec:approach}
RS change conversation aims to describe the semantic changes between the bi-temporal satellite images. 
Another objective is to count the number of change regions present within the scene. 
Additionally, it can focus on explaining the type of changes occurred at region level. However, due to the unavailability of region level ground truth masks, the proposed CDChat currently focus on former two tasks only.

\subsection{CDChat Architecture}
The proposed CDChat utilizes the LLaVA-1.5 \citep{liu2023improvedllava} as the base architecture. As shown in Figure~\ref{fig:cdchat_architecture}, it comprises of three main components, 1) a shared vision encoder for processing the bi-temporal images, 2) a two layer MLP connector, and 3) a large language model (LLM).
Unlike GeoChat and LLaVA, we utilize the Siamese vision encoder to separately extract features from the pre and post-change images, and concatenate these features at the embedding dimension. We then utilize the MLP connector to focus on the change regions and project the features onto the language space which are fed to the language model. Specifically, this approach allows the model to better align the image features with change descriptions thereby improving the model conversational ability. Next, we briefly explain each component of the CDChat.

\noindent\textbf{Vision Encoder: } We utilize the pre-trained vision encoder of CLIP ViT-L-14 \citep{radford2021_clip} for image feature extraction. The encoder is shared as it separately extracts the features of bi-temporal images. Similar to GeoChat \citep{kuckreja2023geochat}, we increase the spatial resolution of RS images to $448\times448$ pixels and correspondingly interpolate the position embedding of the CLIP encoder. This increase in resolution allows the model to pay attention to the small change regions.

\noindent\textbf{MLP Connector: } The MLP connector consists of two linear layers with a GELU activation between them. It takes the concatenated image features of dimension $\mathbb{R}^{1024\times2048}$ and projects them to the language space dimension.

\noindent\textbf{Language Model: } Similar to the LLaVA \citep{liu2023improvedllava} and GeoChat \citep{kuckreja2023geochat}, we utilize the Vicuna-v1.5(7B) \citep{vicuna2023} as the language decoder that takes the text embedding features and output of MLP connector as inputs and generates the text responses to the multimodal prompts. We use the LoRA \citep{hu2022lora} strategy to finetune the language model in order to secure the faster training and enable the model to learn new knowledge without forgetting the previous one.

\subsection{Training Details}
We load the pre-train weights of Vicuna-v1.5 and initialize the vision encoder with CLIP ViT-L-14 \citep{radford2021_clip} weights. We train the model in two stages. First, we freeze the vision encoder and language model and only finetune the MLP connector. Afterwards, we load the weights of tuned MLP connector and freeze it. Then, we use LoRA \citep{hu2022lora} approach to finetune the LLM with a rank of 64 in our implementation.

\begin{table}[t!]
  \centering
  \caption{List of datasets utilized in the generation of instruction file for CDChat. The reported change regions are calculated from the segmentation masks of the respective datasets.}
  \label{tab:dataset_stats}
  \resizebox{\columnwidth}{!}{
  \begin{tabular}{l|cccc}
    \hline
    \textbf{Dataset} & \textbf{Split} & \textbf{\# Image Pairs} & \textbf{\# Change Regions} & \textbf{Image Size}\\
    \hline
    \multirow{2}{*}{LEVIR-CD \citep{chen2020_levircd}}
     & train + val & 3456 & 28819 & \multirow{2}{*}{$256 \times 256$} \\
     & test & 1827 & 8332 &  \\
     \hline
    \multirow{2}{*}{SYSU-CD \citep{shi21sysucd}}
    & train + val & 15665 & 21428 & \multirow{2}{*}{$256 \times 256$} \\
    & test & 3774 & 5396 &  \\
    \hline
  \end{tabular}
  }
\end{table}

\section{Empirical Evaluation}
\subsection{Implementation Details} 
We utilize three Nvidia A100 GPUs to train the model. We finetune the MLP connector for one epoch while LLM and vision encoder are frozen. Afterwards, we LoRA finetune the LLM for one epoch. We utilize the image size of $448 \times 448$ pixels throughout the training and set the batch size equal to 16 per GPU. We use AdamW optimizer with cosine scheduler during training.
\subsection{Datasets} In our experiments, two CD datasets are utilized including LEVIR-CD and SYSU-CD. \textbf{LEVIR-CD} \citep{chen2020_levircd} comprises of 7120, 1024 and 2048 satellite image pairs in train, validation and test sets respectively, having spatial resolution of $256 \times 256$ pixels.
Almost half of the image pairs in the dataset does not contain any changes. Therefore, we remove the image pairs having no changes from the corresponding sets. Remaining image pairs and its change descriptions from train and validation sets are utilized for the instruction data generation.
\textbf{SYSU-CD} \citep{shi21sysucd} contains 12000, 4000, and 4000 image pairs in train, validation and test sets respectively. Each image has a spatial resolution of $256 \times 256$ pixels. Few images in the dataset contain change regions whose change type could not be determined resulting in ambiguous descriptions. Therefore, such images are removed from the respective sets while remaining images and text pairs from train and validation sets are utilized for training. We report the evaluation results on the test sets of the two datasets.
Table~\ref{tab:dataset_stats} shows the statistics of the two datasets listing the number of change regions and image pairs.

\begin{table}[t!]
  \centering
  \caption{Results of change description task on the test set of SYSU-CD.}
  \label{tab:sysu_results}
  \resizebox{\columnwidth}{!}{
  \begin{tabular}{l|cc}
    \hline
    \textbf{Model} & \textbf{METEOR} $(\%) \uparrow$ & \textbf{ROUGE-L} $(\%) \uparrow$\\
    \hline
    MiniGPT-4 \citep{zhu2024minigpt} & 10.94 & 13.48 \\
    LLaVA-1.5 \citep{liu2023improvedllava} & 13.07 & 14.73 \\
    GeoChat \citep{kuckreja2023geochat} & 12.88 & 14.39 \\
    LLaVA++ \citep{hanoona2024LLaVA++} & 13.21 & 13.40 \\
    Gemini-1.5-pro \citep{geminiteam2024gemini} & 14.53 & 14.36 \\
    \rowcolor{Gray}
    CDChat & \textbf{28.27} & \textbf{34.42} \\
    \hline
  \end{tabular}
  }
\end{table}

\subsection{Change Description Task}
We evaluate the performance of the CDChat on the test sets of SYSU-CD \citep{shi21sysucd} and LEVIR-CD \citep{chen2020_levircd} datasets. We provide the model the input image pair and ask question to describe the changes between the two images. The response of the model is recorded for all the image pairs in the test sets. We utilize the METEOR \citep{banerjee2005meteor} and ROUGE-L \citep{lin2004rouge} scores to measure the similarity of the generated response from the model and the annotated change descriptions.

\noindent\textbf{Results: } Table~\ref{tab:levir_results} and \ref{tab:sysu_results} show the performance of various LMMs on LEVIR-CD \citep{chen2020_levircd} and SYSU-CD \citep{chen2020_levircd} respectively.
On SYSU-CD, LLaVA-1.5 \citep{liu2023improvedllava} performs better compared to other LMMs by achieving METEOR \citep{banerjee2005meteor} and ROUGE-L \citep{lin2004rouge} scores of 13.07\% and 14.73\% respectively indicating better generalization abilities. Even though, GeoChat \citep{kuckreja2023geochat} surpasses all these models in RS scene classification, RS image and region grounding tasks, however, it performance degraded for RS change description task. Notably, CDChat outperforms all the LMMs and achieves ROUGE-L \citep{lin2004rouge} score of 34.42\%.
In case of LEVIR-CD, multiple ground truth change descriptions are available for each image pair. Therefore, the scores are computed by utilizing multiple ground truth references. From Table.~\ref{tab:levir_results}, we observe that the performance trend of the models are similar to that of SYSU-CD. The performance of LLaVA-1.5 is better than the other LMMs by achieving METEOR score of 23.74\%. However, CDChat performs significantly better as compared to the listed LMMs.

\begin{table}[t!]
  \centering
  \caption{Results of change description task on the test set of LEVIR-CD.}
  \label{tab:levir_results}
  \resizebox{\columnwidth}{!}{
  \begin{tabular}{l|cc}
    \hline
    \textbf{Model} & \textbf{METEOR} $(\%) \uparrow$ & \textbf{ROUGE-L} $(\%) \uparrow$\\
    \hline
    MiniGPT-4 \citep{zhu2024minigpt} & 15.62 & 13.10 \\
    LLaVA-1.5 \citep{liu2023improvedllava} & 23.74 & 15.37 \\
    GeoChat \citep{kuckreja2023geochat} & 21.29 & 14.56 \\
    LLaVA++ \citep{hanoona2024LLaVA++} & 21.75 & 12.87 \\
    Gemini-1.5-pro \citep{geminiteam2024gemini} & 22.59 & 13.76 \\
    \rowcolor{Gray}
    CDChat & \textbf{36.39} & \textbf{23.86} \\
    \hline
  \end{tabular}
  }
\end{table}

\subsection{Change Region Counting} 
In this task, we provide the LMM the pair of bi-temporal images and ask it to provide the count or number of change regions. Here, count is a range of intervals and LMM has to choose the answer from one of those intervals. Specifically, we ask following type of question to the LMM:

\textit{
How many change regions are there in the two images? Choose from the given ranges: less than or equal to five, between six and ten, between eleven and twenty, more than twenty.}

The responses from each LMM listed in Table.~\ref{tab:sysu_results} are saved in the files and accuracy score is computed. We observe that all models are unable to answer the counting questions despite that the instructions are given in the question. However, our CDChat performs reasonably and provide accuracy score of 68.97\% and 83.25\% on SYSU-CD and LEVIR-CD test sets respectively.

\section{Conclusion}
\label{sec:conclusion}
In this study, we propose a CDChat for describing the changes between the RS images. We conclude that the existing LMMs strive to explain the changes between the RS images. Therefore, an explicit instruction dataset is required for the LMM to improve performance. 
We also infer that the existing LMMs are unable to generate response to the type of question that ask to choose a range from the given options and the LMM has to be explicitly learn these type of examples.

Our potential future direction is to extend the capabilities of CDChat to incorporate series of satellite images and multilinguality.

\section{Limitations}
\label{sec:limitations}
CDChat requires image pair as an input to the LMM which limits its ability to perform only change description task. Due to this limitation, CDChat cannot be utilized for image or region level grounding task or classify an image like GeoChat or LLaVA-1.5. Additionally, lack of change description datasets restricts the generalization performance of CDChat. 
As discussed in section.~\ref{sec:conclusion}, a potential future direction is to extend the functionality of CDChat to incorporate series of satellite images and support multi-sensor RS images.

\bibliography{custom}

\end{document}